\documentclass{article} 
\usepackage{icais2025_conference,times}
\iclrfinalcopy

\usepackage{amsmath,amsfonts,bm}

\def\eqref#1{equation~\ref{#1}}

\def\1{\bm{1}}

\DeclareMathAlphabet{\mathsfit}{\encodingdefault}{\sfdefault}{m}{sl}
\SetMathAlphabet{\mathsfit}{bold}{\encodingdefault}{\sfdefault}{bx}{n}

\usepackage{hyperref}
\hypersetup{
    colorlinks=true,
    linkcolor=blue!70!black,
    citecolor=green!60!black,
    urlcolor=red!60!black
}
\usepackage{url}
\usepackage{graphicx}

\title{LLM-empowered knowledge graph construction: A survey}

\author{Haonan Bian \\
Xidian University\\
Xi'an China \\
\texttt{23151214251@stu.xidian.edu.cn}}

\begin{document}

\maketitle

\begin{abstract}
Knowledge Graphs (KGs) have long served as a fundamental infrastructure for structured knowledge representation and reasoning. With the advent of Large Language Models (LLMs), the construction of KGs has entered a new paradigm—shifting from rule-based and statistical pipelines to language-driven and generative frameworks. This survey provides a comprehensive overview of recent progress in \textbf{LLM-empowered knowledge graph construction}, systematically analyzing how LLMs reshape the classical three-layered pipeline of ontology engineering, knowledge extraction, and knowledge fusion.

We first revisit traditional KG methodologies to establish conceptual foundations, and then review emerging LLM-driven approaches from two complementary perspectives: \textit{schema-based} paradigms, which emphasize structure, normalization, and consistency; and \textit{schema-free} paradigms, which highlight flexibility, adaptability, and open discovery. Across each stage, we synthesize representative frameworks, analyze their technical mechanisms, and identify their limitations.

Finally, the survey outlines key trends and future research directions, including KG-based reasoning for LLMs, dynamic knowledge memory for agentic systems, and multimodal KG construction. Through this systematic review, we aim to clarify the evolving interplay between LLMs and knowledge graphs, bridging symbolic knowledge engineering and neural semantic understanding toward the development of adaptive, explainable, and intelligent knowledge systems.
\end{abstract}

\section{Introduction}

Knowledge Graphs (KGs) have long served as a cornerstone for representing, integrating, and reasoning over structured knowledge. They provide a unified semantic foundation that underpins a wide range of intelligent applications, such as semantic search, question answering, and scientific discovery. Conventional KG construction pipelines are typically composed of three major components: \textbf{ontology engineering}, \textbf{knowledge extraction}, and \textbf{knowledge fusion}. Despite their success in enabling large-scale knowledge organization, traditional paradigms (e.g.,~\citet{zhong2023comprehensive};~\citet{zhao2024survey}) continue to face three enduring challenges: (1) \textit{Scalability and data sparsity}, as rule-based and supervised systems often fail to generalize across domains; (2) \textit{Expert dependency and rigidity}, since schema and ontology design require substantial human intervention and lack adaptability; and (3) \textit{Pipeline fragmentation}, where the disjoint handling of construction stages causes cumulative error propagation. These limitations hinder the development of self-evolving, large-scale, and dynamic KGs.

The advent of \textbf{Large Language Models (LLMs)} introduces a transformative paradigm for overcoming these bottlenecks. Through large-scale pretraining and emergent generalization capabilities, LLMs enable three key mechanisms: (1) \textit{Generative knowledge modeling}, synthesizing structured representations directly from unstructured text; (2) \textit{Semantic unification}, integrating heterogeneous knowledge sources through natural language grounding; and (3) \textit{Instruction-driven orchestration}, coordinating complex KG construction workflows via prompt-based interaction. Consequently, LLMs are evolving beyond traditional text-processing tools into \textit{cognitive engines} that seamlessly bridge natural language and structured knowledge (e.g.,~\citet{zhu2024llms};~\citet{zhangExtractDefineCanonicalize2024}). This evolution marks a paradigm shift from \textit{rule-driven, pipeline-based systems} toward \textit{LLM-driven, unified, and adaptive frameworks}, where knowledge acquisition, organization, and reasoning emerge as interdependent processes within a generative and self-refining ecosystem~\citep{pan2024unifying}.

In light of these rapid advances, this paper presents a comprehensive \textbf{survey} of LLM-driven knowledge graph construction. We systematically review recent research spanning ontology engineering, knowledge extraction, and fusion, analyze emerging methodological paradigms, and highlight open challenges and future directions at the intersection of LLMs and knowledge representation.

The remainder of this paper is organized as follows:
\begin{itemize}
    \item \textbf{Section 2} introduces the foundations of traditional knowledge graph construction, covering ontology engineering, knowledge extraction, and fusion techniques prior to the LLM era.  
    \item \textbf{Section 3} reviews LLM-enhanced ontology construction, encompassing both top-down paradigms (LLMs as ontology assistants) and bottom-up paradigms (KGs for LLMs).  
    \item \textbf{Section 4} presents LLM-driven knowledge extraction, comparing schema-based and schema-free methodologies.  
    \item \textbf{Section 5} discusses LLM-powered knowledge fusion, focusing on schema-level, instance-level, and hybrid frameworks.  
    \item \textbf{Section 6} explores future research directions, including KG-based reasoning, dynamic knowledge memory, and multimodal KG construction.  
\end{itemize}

\section{Preliminaries}

The construction of Knowledge Graphs (KGs) traditionally follows a \textbf{three-layered pipeline} comprising \textit{ontology engineering}, \textit{knowledge extraction}, and \textit{knowledge fusion}. Prior to the advent of Large Language Models (LLMs), these stages were implemented through rule-based, statistical, and symbolic approaches. This section briefly reviews these conventional methodologies to establish context for the subsequent discussion on LLM-empowered KG construction.

\subsection{Ontology Engineering}

Ontology Engineering (OE) involves the formal specification of domain concepts, relationships, and constraints. In the pre-LLM era, ontologies were primarily \textbf{manually constructed by domain experts}, often supported by semantic web tools such as \textit{Protégé} and guided by established methodologies including \textit{METHONTOLOGY} and \textit{On-To-Knowledge}. These systematic processes emphasized conceptual rigor and logical consistency but required extensive expert intervention.  

As summarized by \citet{zouaqSurveyDomainOntology2010}, ontology design during this period was characterized by strong human supervision and limited scalability. Subsequent semi-automatic approaches—collectively known as \textit{ontology learning}—sought to derive ontological structures from textual corpora, as reviewed in \citet{asim2018survey}. However, even advanced frameworks such as \textit{NeOn} struggled with ontology evolution, modular reuse, and dynamic adaptation. As highlighted by \citet{Kotis_Vouros_Spiliotopoulos_2020}, traditional OE frameworks offered precision and formal soundness but lacked flexibility and efficiency for large-scale or continuously evolving knowledge domains.

\subsection{Knowledge Extraction}

Knowledge Extraction (KE) aims to identify \textbf{entities, relations, and attributes} from unstructured or semi-structured data. Early approaches relied on handcrafted linguistic rules and pattern matching, which provided interpretability but were brittle and domain-specific. The evolution from symbolic and rule-based systems to statistical and neural methods has been systematically summarized in \citet{pai-etal-2024-survey}.  

The advent of deep learning architectures, such as \textit{BiLSTM-CRF} and \textit{Transformer}-based models, marked a paradigm shift toward data-driven feature learning, as discussed by \citet{yang2022survey}. Comprehensive analyses such as \citet{detroja2023survey} further categorize supervised, weakly supervised, and unsupervised relation extraction paradigms, emphasizing their dependence on annotated data and limited cross-domain generalization.  

In summary, traditional KE methods established the technical foundation for modern extraction pipelines but remained constrained by \textbf{data scarcity, weak generalization, and cumulative error propagation}—limitations that motivate the LLM-driven paradigms discussed in later sections.

\subsection{Knowledge Fusion}

Knowledge Fusion (KF) focuses on integrating heterogeneous knowledge sources into a coherent and consistent graph by resolving issues of duplication, conflict, and heterogeneity. A central subtask is \textbf{entity alignment}, which determines whether entities from different datasets refer to the same real-world object.  

Classical approaches relied on lexical and structural similarity measures, as reviewed in \citet{ZENG20211}. The introduction of representation learning enabled embedding-based techniques that align entities within shared vector spaces, improving scalability and automation, as surveyed by \citet{zhu2024survey}.  
Domain-specific applications, such as \citet{Yang2022_Collective}, demonstrate multi-feature fusion strategies combining structural, attribute, and relational similarities. Other graph-level models, such as \citet{app12199434}, further integrate semantic cues to enhance alignment robustness.  

Despite these advancements, traditional fusion pipelines continue to struggle with \textbf{semantic heterogeneity, large-scale integration, and dynamic knowledge updating}—challenges that contemporary LLM-based fusion frameworks are increasingly designed to address.

\begin{figure}[h]
    \centering
    \includegraphics[width=0.95\linewidth]{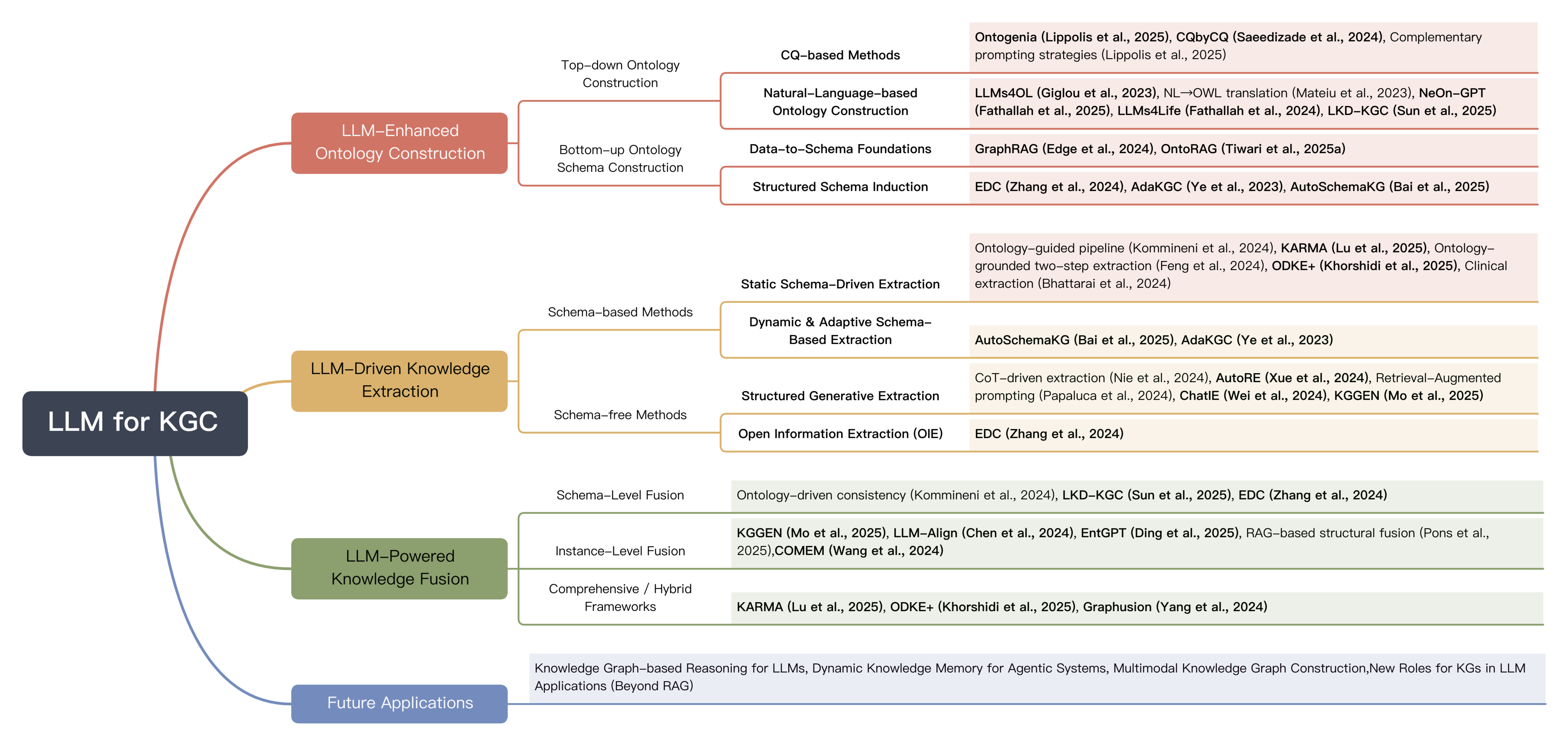}
    \caption{Taxonomy of LLM for KGC}
    \label{fig:placeholder}
\end{figure}

\section{LLM-Enhanced Ontology Construction}
The integration of Large Language Models (LLMs) has introduced a fundamental paradigm shift in Ontology Engineering (OE) and, by extension, Knowledge Graph (KG) construction. Current research generally follows two complementary directions: a \textbf{top-down} approach, which leverages LLMs as intelligent assistants for formal ontology modeling, and a \textbf{bottom-up} approach, which employs ontology construction to enhance the reasoning and representation capabilities of LLMs themselves.

\subsection{Top-down Ontology Construction: LLMs as Ontology Assistants}
The top-down paradigm extends the traditions of the Semantic Web and Knowledge Engineering, emphasizing ontology development guided by predefined semantic requirements. Within this framework, LLMs serve as advanced co-modelers that assist human experts in translating natural language specifications---such as competency questions (CQs), user stories, or domain descriptions---into formal ontologies, typically represented in OWL or related standards. This paradigm prioritizes conceptual abstraction, the precise definition of relations, and structured semantic representation to ensure that subsequent knowledge extraction and instance population adhere to well-defined logical constraints.

\subsubsection{Competency Question (CQ)-Based Ontology Construction}
CQ-based methods represent a requirements-driven pathway toward automated ontology modeling. In this setting, LLMs parse CQs or user stories to identify, categorize, and formalize domain-specific concepts, attributes, and relationships.

A pioneering framework, \textbf{Ontogenia}~\citep{lippolisOntogeniaOntologyGeneration2025}, introduced the use of Metacognitive Prompting for ontology generation, enabling the model to perform self-reflection and structural correction during synthesis. By incorporating Ontology Design Patterns (ODPs), Ontogenia improves both the consistency and complexity of generated ontologies. Similarly, the \textbf{CQbyCQ} framework~\citep{saeedizadeNavigatingOntologyDevelopment2024} demonstrated that LLMs can directly translate CQs and user stories into OWL-compliant schemas, effectively automating the transition from requirements to structured ontological models.

Building on these advances, \citet{lippolisOntologyGenerationUsing2025} proposed two complementary prompting strategies: a ``memoryless'' approach for modular construction and a reflective iterative method inspired by Ontogenia. Empirical evaluations revealed that LLMs can autonomously identify classes, object properties, and data properties, while generating corresponding logical axioms with consistency comparable to that of junior human modelers. Collectively, these studies have led to semi-automated ontology construction pipelines encompassing the entire lifecycle---from CQ formulation and validation to ontology instantiation---with human experts intervening only at critical checkpoints. Through this evolution, LLMs have transitioned from passive analytical tools to active modeling collaborators in ontology design.

\subsubsection{Natural Language-Based Ontology Construction}
Beyond CQ-driven paradigms, natural language-based ontology construction seeks to induce semantic schemas directly from unstructured or semi-structured text corpora, eliminating the dependency on explicitly formulated questions. The goal is to enable LLMs to autonomously uncover conceptual hierarchies and relational patterns from natural language, achieving a direct mapping from textual descriptions to formal logical representations.

Foundational work in this domain---including \citet{saeedizadeNavigatingOntologyDevelopment2024} and \citet{lippolisOntologyGenerationUsing2025}---systematically evaluated GPT-4’s performance and confirmed that its outputs approach the quality of novice human modelers, thereby validating the feasibility of ``intelligent ontology assistants.'' The \textbf{LLMs4OL} framework~\citep{giglouLLMs4OLLargeLanguage2023} further verified LLMs’ capacity for concept identification, relation extraction, and semantic pattern induction in general-purpose domains. Likewise, \citet{mateiuOntologyEngineeringLarge2023} demonstrated the use of fine-tuned models to directly translate natural language into OWL axioms within established ontology editors such as Protégé.

Recent systems such as \textbf{NeOn-GPT}~\citep{fathallahNeOnGPTLargeLanguage2025} and \textbf{LLMs4Life}~\citep{fathallahLLMs4LifeLargeLanguage2024} have advanced this direction by introducing end-to-end, prompt-driven workflows that integrate ontology reuse and adaptive refinement to construct deep, coherent ontological structures in complex scientific domains (e.g., life sciences). Meanwhile, lightweight frameworks such as \textbf{LKD-KGC}~\citep{sunLKDKGCDomainSpecificKG2025} enable rapid schema induction for open-domain knowledge graphs by clustering entity types extracted from document summaries. 

In summary, top-down research on LLM-assisted ontology construction emphasizes semantic consistency, structural completeness, and human–AI collaboration, marking a significant evolution of traditional knowledge engineering toward more intelligent, language-driven paradigms.

\subsection{Bottom-up Ontology Schema Construction: KGs for LLMs}

The bottom-up methodology has gained increasing attention as a response to paradigm shifts introduced by the era of Large Language Models (LLMs), particularly within Retrieval-Augmented Generation (RAG) frameworks. In this paradigm, the knowledge graph is no longer viewed merely as a static repository of structured knowledge for human interpretation. Instead, it serves as a dynamic infrastructure that provides factual grounding and structured memory for LLMs. Consequently, research focus has shifted from manually designing ontological hierarchies to \textbf{automatically inducing schemas} from unstructured or semi-structured data. This evolution can be delineated through three interrelated stages of progress.

Early studies such as \textbf{GraphRAG}~\citep{edgeLocalGlobalGraph2024} and \textbf{OntoRAG}~\citep{tiwariOntoRAGEnhancingQuestionAnswering2025a} established the foundation for data-driven ontology construction. These approaches first generate instance-level graphs from raw text via open information extraction, and then abstract ontological concepts and relations through clustering and generalization. This ``data-to-schema'' process transforms empirical knowledge into reusable conceptual structures, illustrating how instance-rich corpora can give rise to ontological blueprints.

Building upon this foundation, the \textbf{EDC (Extract–Define–Canonicalize)} framework~\citep{zhangExtractDefineCanonicalize2024} advanced the pipeline into a three-stage process consisting of open extraction, semantic definition, and schema normalization. It enables the alignment of automatically induced schemas with existing ontologies, or the creation of new ones when predefined structures are absent. Extending this adaptability, \textbf{AdaKGC}~\citep{yeSchemaadaptableKnowledgeGraph2023} addressed the challenge of dynamic schema evolution, allowing models to incorporate novel relations and entity types without retraining. Collectively, these advances shift the focus from static schema construction toward \textbf{continuous schema adaptation} within evolving knowledge environments.

More recent efforts have transitioned beyond algorithmic prototypes toward deployable knowledge systems. For example, \textbf{AutoSchemaKG}~\citep{baiAutoSchemaKGAutonomousKnowledge2025} integrates schema-based and schema-free paradigms within a unified architecture, supporting the real-time generation and evolution of enterprise-scale knowledge graphs. In this stage, KGs operate as a form of \textit{external knowledge memory} for LLMs—prioritizing factual coverage, scalability, and maintainability over purely semantic completeness. This transformation marks a pragmatic reorientation of ontology construction, emphasizing its service to LLM reasoning and interpretability in knowledge-intensive applications.

In summary, bottom-up ontology schema construction redefines the interplay between LLMs and knowledge engineering. The focus evolves from \textbf{``LLMs for Ontology Engineering''} to \textbf{``Ontologies and KGs for LLMs''}. Whereas the \textit{top-down} trajectory emphasizes semantic modeling, logical consistency, and expert-guided alignment—positioning LLMs as intelligent assistants in ontology design—the \textit{bottom-up} trajectory prioritizes automatic extraction, schema induction, and dynamic evolution. This progression advances toward self-updating, interpretable, and scalable knowledge ecosystems that strengthen the grounding and reasoning capabilities of LLMs.

\section{LLM-Driven Knowledge Extraction}

Through a systematic examination of recent advances, it becomes evident that methodologies for Large Language Model (LLM)-driven knowledge extraction have evolved along two principal paradigms: \textbf{schema-based} extraction, which operates under explicit structural guidance, and \textbf{schema-free} extraction, which transcends the limitations of predefined templates. The former emphasizes normalization, structural consistency, and semantic alignment, while the latter prioritizes adaptability, openness, and exploratory discovery. Together, these paradigms delineate the conceptual landscape of contemporary research in LLM-based knowledge extraction.

\subsection{Schema-Based Methods}

The central principle of schema-based extraction lies in its reliance on an explicit knowledge schema that provides both structural guidance and semantic constraints for the extraction process. Within this paradigm, the research trajectory demonstrates a clear evolution—from the use of static ontological blueprints toward adaptive and dynamically evolving schema frameworks.

\subsubsection{Static Schema-Driven Extraction}

Early studies in LLM-driven knowledge extraction predominantly employed \textbf{predefined, static schemas} that rigidly constrained the extraction process. In this paradigm, the ontology functions as a fixed semantic backbone, directing the LLM to populate the knowledge base under strict structural supervision. The progression of this research line can be broadly characterized by three developmental stages.

Initial efforts, such as \citet{kommineni2024towards}, utilized fully predefined ontological structures to ensure precision and interpretability. In their pipeline, the LLM first generates Competency Questions (CQs) to delineate the knowledge scope, constructs the corresponding ontology (TBox), and subsequently performs ABox population under explicit schema supervision—achieving high consistency but limited flexibility. Similarly, the \textbf{KARMA} framework~\citep{luKARMALeveragingMultiAgent2025} adopts a multi-agent architecture, in which each agent executes schema-guided extraction tasks to guarantee accurate entity normalization and relation classification within a fixed ontological boundary.

Building on these rigid frameworks, subsequent studies sought to enhance modularity and reusability through staged prompting. For instance, \citet{fengOntologygroundedAutomaticKnowledge2024} proposed a two-step ``ontology-grounded extraction'' approach: first generating a domain-specific ontology directly from text, and then leveraging it as a directive prompt for RDF triple extraction. This approach strengthens structural alignment while maintaining partial adaptability. 

More recent developments introduce localized flexibility within otherwise static frameworks. \textbf{ODKE+}~\citep{khorshidiODKEOntologyGuidedOpenDomain2025} proposes \textit{ontology snippets}—dynamically selected ontology subsets—to construct context-aware prompts tailored to specific entities, thus enabling limited schema adaptation at runtime. Likewise, \citet{bhattaraiDocumentlevelClinicalEntity2024} utilize the medical ontology UMLS to dynamically generate task-specific prompts for clinical information extraction. Although these methods introduce local adaptability, they remain bounded by pre-existing macro-schemas, representing a transitional phase toward schema dynamism.

In summary, static schema-driven extraction forms the foundational paradigm of LLM-assisted knowledge extraction, emphasizing \textbf{precision, logical consistency, and interpretability}. However, its dependence on rigid ontological templates restricts scalability and cross-domain generalization. The progression from fixed schema control to selective, context-aware schema prompting marks the field’s gradual shift toward more adaptive, data-responsive frameworks.

\subsubsection{Dynamic and Adaptive Schema-Based Extraction}

Recent approaches reconceptualize the schema as a \textbf{dynamic, evolving component} of the extraction process rather than a fixed template—a shift from schema \textit{guiding} extraction to schema \textit{co-evolving} with it.

\textbf{AutoSchemaKG}~\citep{baiAutoSchemaKGAutonomousKnowledge2025} exemplifies this trend by inducing schemas from large-scale corpora via unsupervised clustering and relation discovery. It employs multi-stage prompts tailored to different relation types, enabling the schema to evolve iteratively with extracted content and improving open-domain scalability. Building on this idea, \textbf{AdaKGC}~\citep{yeSchemaadaptableKnowledgeGraph2023} tackles \textit{schema drift} through two mechanisms: the \textbf{Schema-Enriched Prefix Instruction (SPI)} for context-aware prompting and the \textbf{Schema-Constrained Dynamic Decoding (SDD)} for schema adaptation without retraining. 

Together, these methods enable \textbf{adaptive schema learning}, bridging symbolic structure and data-driven flexibility. They lay the groundwork for continual, self-updating knowledge graph construction where extraction and schema evolution progress synergistically.

\subsection{Schema-free Methods}

In contrast to paradigms that depend on externally defined blueprints, \textbf{schema-free extraction} aims to acquire structured knowledge directly from unstructured text without relying on any predefined ontology or relation schema. The central idea is to leverage Large Language Models (LLMs) as autonomous extractors capable of identifying entities and relations through advanced prompt engineering, instruction tuning, and self-organizing reasoning. The evolution of this paradigm unfolds along two major trajectories: \textit{structured generative extraction} and \textit{open information extraction}.

\subsubsection{Structured Generative Extraction}

The first trajectory, \textbf{structured generative extraction}, focuses on prompting LLMs to construct an \textit{implicit} or \textit{on-the-fly} schema during generation. Although no external ontology is provided, structured reasoning patterns and generative templates guide the model toward consistent and coherent knowledge generation.

Early studies, such as \citet{nie2024knowledge}, integrated the extraction process with \textbf{Chain-of-Thought (CoT)} prompting, encouraging stepwise reasoning for entity and relation identification. This approach demonstrated that explicit schemas can be effectively replaced by reasoning-driven organization. Building on this insight, \textbf{AutoRE}~\citep{xueAutoREDocumentLevelRelation2024} introduced an RHF (Relation–Head–Facts) pipeline via instruction fine-tuning, enabling the model to internalize relational regularities and improve coherence and scalability across documents.

Subsequent works further enhanced schema-free extraction by incorporating retrieval and interactivity. For instance, \citet{papalucaZeroFewShotsKnowledge2024} proposed a \textbf{Retrieval-Augmented} prompting framework that dynamically enriches the context window with semantically related exemplars, thereby improving factual precision. In parallel, \textbf{ChatIE}~\citep{weiChatIEZeroShotInformation2024} reformulated extraction as a \textbf{multi-turn dialogue} process, wherein the model iteratively refines entity and relation candidates through chained question answering. Similarly, \textbf{KGGEN}~\citep{moKGGenExtractingKnowledge2025} decomposed extraction into two sequential LLM invocations—first detecting entities, then generating relations—to reduce cognitive load and mitigate error propagation.

Collectively, these studies reveal that even without explicit schemas, LLMs can internalize latent relational structures through guided reasoning, modular prompting, and interactive refinement—laying the groundwork for open-ended and self-organizing knowledge generation.

\subsubsection{Open Information Extraction (OIE)}

\textbf{Open Information Extraction (OIE)} extends structured generative methods toward schema-free extraction, aiming to discover all possible entity–relation–object triples from text without relying on predefined types.  The \textbf{EDC} framework~\citep{zhangExtractDefineCanonicalize2024} exemplifies this paradigm: its \textit{Extract} stage uses few-shot prompting to generate comprehensive natural-language triples, producing a raw open knowledge graph that later undergoes definition and canonicalization.  OIE prioritizes coverage and discovery over structural regularity. When combined with schema induction or canonicalization, it bridges unstructured text and emergent ontological organization—completing the continuum from schema-free to schema-generative knowledge construction.

\section{LLM-Powered Knowledge Fusion}

An examination of recent advances and pioneering studies indicates that methodologies leveraging Large Language Models (LLMs) for knowledge fusion predominantly address challenges at two fundamental levels: (1) constructing a unified and normalized knowledge \textit{skeleton} at the \textbf{schema layer}, and (2) integrating and aligning the specific knowledge \textit{flesh} at the \textbf{instance layer}. According to this distinction, existing approaches can be categorized into three major classes: \textbf{schema-level fusion}, \textbf{instance-level fusion}, and \textbf{hybrid frameworks} that integrate both.

\subsection{Schema-Level Fusion}

Schema-level fusion unifies the \textbf{structural backbone} of knowledge graphs—concepts, entity types, relations, and attributes—into a coherent and semantically consistent schema. By aligning heterogeneous elements, it ensures that all knowledge adheres to a unified conceptual specification. Research in this area has evolved through three major phases.

\textbf{(1) Ontology-driven consistency.}  
Early work relied on explicit ontologies as global constraints. For instance, \citet{kommineni2024towards} enforced alignment between extracted triples and predefined ontological definitions, achieving high semantic consistency but limited flexibility across domains.

\textbf{(2) Data-driven unification.}  
To overcome this rigidity, \textbf{LKD-KGC}~\citep{sunLKDKGCDomainSpecificKG2025} introduced adaptive, embedding-based schema integration. It automatically extracts and merges equivalent entity types via \textit{vector clustering} and \textit{LLM-based deduplication}, allowing schema alignment to emerge dynamically from data.

\textbf{(3) LLM-enabled canonicalization.}  
Recent approaches such as \textbf{EDC}~\citep{zhangExtractDefineCanonicalize2024} extend fusion toward semantic canonicalization. By prompting LLMs to generate natural language definitions of schema components and comparing them via vector similarity, this method supports both self-alignment and cross-schema mapping with greater automation and semantic precision.

In summary, schema-level fusion has progressed from \textbf{ontology-driven} to \textbf{data-driven} to \textbf{LLM-enabled} paradigms—marking a shift from rigid rule-based alignment toward flexible, semantics-oriented fusion mediated by LLM reasoning.

\subsection{Instance-Level Fusion}

Instance-level fusion integrates concrete knowledge instances by addressing \textbf{entity alignment, disambiguation, deduplication}, and \textbf{conflict resolution}. Its goal is to reconcile heterogeneous or redundant entities to maintain a coherent and semantically precise knowledge graph. Recent work reflects a clear evolution—from heuristic clustering to structured, reasoning-based frameworks.

Early studies such as \textit{KGGEN}~\citep{moKGGenExtractingKnowledge2025} employed iterative LLM-guided clustering to merge equivalent entities and relations beyond surface matching. The framework performs progressive triple extraction followed by semantic grouping, revealing the potential of LLMs to aggregate semantically related entities through implicit reasoning rather than explicit rules.  
Later, \textit{LLM-Align}~\citep{chenLLMAlignUtilizingLarge2024} and \textit{EntGPT}~\citep{dingEntGPTEntityLinking2025} reframed alignment as a \textit{contextual reasoning} task, using multi-step prompting to enhance semantic discrimination. \textit{LLM-Align} treats alignment as a constrained multiple-choice problem, while \textit{EntGPT} introduces a two-phase refinement pipeline that first generates candidate entities and then applies targeted reasoning for final selection, significantly improving alignment precision.  
More recent efforts incorporate structural and retrieval cues—e.g., \citet{ponsKnowledgeGraphsEnhancing2025} leverage \textbf{RAG}-based fusion to exploit class–subclass hierarchies and entity descriptions for zero-shot disambiguation. This integration of graph topology enables more robust reasoning about unseen or ambiguous entities.  
Efficiency has also improved through hierarchical designs such as \textbf{COMEM}~\citep{wang2024match}, which combines lightweight filtering with fine-grained reasoning. By cascading smaller and larger LLMs in a multi-stage pipeline, it achieves substantial efficiency gains while maintaining high semantic accuracy in large-scale fusion tasks.

Overall, LLMs have evolved from simple matchers to \textbf{adaptive reasoning agents} that integrate contextual, structural, and retrieved signals for scalable, self-correcting fusion—paving the way toward autonomous knowledge graph construction.

\subsection{Comprehensive and Hybrid Frameworks}

Comprehensive and hybrid frameworks unify \textbf{schema-level} and \textbf{instance-level} fusion within a single, end-to-end workflow, moving beyond traditional modular pipelines toward integrated, prompt-driven architectures.

The \textit{KARMA} framework~\citep{luKARMALeveragingMultiAgent2025} exemplifies a multi-agent design where specialized agents collaboratively handle schema alignment, conflict resolution, and quality evaluation, achieving scalability and global consistency. Building on this, \textit{ODKE+}~\citep{khorshidiODKEOntologyGuidedOpenDomain2025} employs an ontology-guided workflow coupling schema supervision with instance-level corroboration for improved semantic fidelity. More recently, \textit{Graphusion}~\citep{yangGraphusionLeveragingLarge2024} introduces a unified, prompt-based paradigm that performs all fusion subtasks—alignment, consolidation, and inference—within a single generative cycle.

Together, these frameworks signal a shift toward \textbf{integrated, adaptive, and generative} fusion systems, marking a crucial step toward autonomous, self-evolving knowledge graphs capable of continuous construction and reasoning in LLM-driven ecosystems.

\section{Future Applications}

Research at the intersection of Large Language Models (LLMs) and Knowledge Graphs (KGs) is progressively advancing toward deeper intelligent interaction and greater autonomy in knowledge representation and reasoning. In light of these developments, several promising directions are emerging for future exploration.

\subsection{Knowledge Graph-based Reasoning for LLMs}

Future work is expected to further integrate structured KGs into the reasoning mechanisms of LLMs, enhancing their logical consistency, causal inference, and interpretability. This research direction signifies not only an improvement in reasoning capabilities but also a conceptual transition from \emph{knowledge construction} to \emph{knowledge-driven reasoning}. High-quality, well-structured KGs will provide the foundation for explainable and verifiable model inference. Existing studies such as \textit{KG-based Random-Walk Reasoning}~\citep{kimCausalReasoningLarge2024} and \textit{KG-RAR}~\citep{wuGraphAugmentedReasoningEvolving2025} have demonstrated the potential of this paradigm. A crucial complementary challenge, however, lies in how enhanced reasoning abilities can in turn support more robust and automated KG construction—forming a self-improving, virtuous cycle between knowledge building and reasoning.

\subsection{Dynamic Knowledge Memory for Agentic Systems}

Achieving \textbf{autonomy} in LLM-powered agents requires overcoming the limits of finite context windows through \textbf{persistent, structured memory}. Recent architectures envision the \textbf{knowledge graph (KG)} as a \textit{dynamic memory substrate}, evolving continuously with agent interactions rather than storing static histories.
Frameworks such as \textbf{A-MEM}~\citep{xuAMEMAgenticMemory2025} model memory as interconnected “notes” enriched with contextual metadata, enabling continual reorganization and growth. Similarly, \textbf{Zep}~\citep{rasmussenZepTemporalKnowledge2025} employs a \textbf{temporal knowledge graph (TKG)} to manage fact validity and support time-aware reasoning and updates.
These advances highlight dynamic KGs as long-term, interpretable memory systems that enable \textbf{continuous learning}, \textbf{multi-agent collaboration}, and \textbf{self-reflective reasoning}. Future work will focus on improving scalability, temporal coherence, and multimodal integration for fully autonomous, knowledge-grounded agents.

\subsection{Multimodal Knowledge Graph Construction}

Multimodal Knowledge Graph (MMKG) construction aims to integrate heterogeneous modalities—such as text, images, audio, and video—into unified, structured representations that enable richer reasoning and cross-modal alignment. Representative work includes \textbf{VaLiK} (Vision-align-to-Language integrated KG)~\citep{liuAligningVisionLanguage2025}, which cascades pretrained Vision-Language Models (VLMs) to translate visual features into textual form, followed by a cross-modal verification module to filter noise and assemble MMKGs. This process achieves entity–image linkage without manual annotation. Beyond structure, representation learning methods such as \textbf{KG-MRI}~\citep{ijcai2024p659} employ multimodal embeddings with contrastive objectives to align heterogeneous modalities into coherent semantic spaces. Key challenges remain in modality heterogeneity, alignment noise, scalability, and robustness under missing or imbalanced modalities. As LLMs and VLMs continue to co-evolve, MMKGs will become a cornerstone for bridging perceptual input and symbolic reasoning across modalities.

\subsection{New Roles for KGs in LLM Applications: Beyond RAG}

Beyond their use as retrieval backbones in RAG systems, Knowledge Graphs (KGs) are increasingly envisioned as a \textbf{cognitive middle layer} bridging raw input and LLM reasoning. In this paradigm, KGs provide a structured scaffold for \textbf{querying, planning, and decision-making}, enabling more interpretable and grounded generation.
Recent studies illustrate this shift. \textbf{CogER}~\citep{bingCognitionawareKnowledgeGraph2023} formulates recommendation as cognition-aware KG reasoning, integrating intuitive and path-based inference for explainability. In the biomedical domain, \textbf{PKG-LLM}~\citep{sarabadaniPKGLLMFrameworkPredicting2025} employs domain KGs for knowledge augmentation and predictive modeling in mental health diagnostics. Together, these approaches treat the KG as an \textbf{interactive reasoning substrate}, promising more robust and explainable generation in domains such as science, code, and healthcare.

\section{Conclusion}

This survey presents a comprehensive overview of how Large Language Models (LLMs) are transforming Knowledge Graph (KG) construction across ontology engineering, knowledge extraction, and knowledge fusion. LLMs shift the paradigm from rule-based and modular pipelines toward unified, adaptive, and generative frameworks.
Across these stages, three trends emerge: (1) the evolution from \textit{static schemas} to \textit{dynamic induction}, (2) the integration of \textit{pipeline modularity} into \textit{generative unification}, and (3) the transition from \textit{symbolic rigidity} to \textit{semantic adaptability}. Together, these shifts redefine KGs as \textbf{living, cognitive infrastructures} that blend language understanding with structured reasoning.
Despite remarkable progress, challenges remain in scalability, reliability, and continual adaptation. Future advances in \textit{prompt design}, \textit{multimodal integration}, and \textit{knowledge-grounded reasoning} will be key to realizing \textbf{autonomous and explainable knowledge-centric AI systems}.

\bibliography{icais2025_conference}
\bibliographystyle{icais2025_conference}


\end{document}